\documentclass{article} 
\usepackage[final]{colm2026_conference}

\usepackage[T1]{fontenc}
\usepackage{microtype}
\usepackage{hyperref}
\usepackage{url}
\usepackage[normalem]{ulem}
\usepackage{booktabs}
\usepackage{multirow}
\usepackage{fontawesome5}
\usepackage{graphicx}
\usepackage{xcolor}
\usepackage[most]{tcolorbox}
\usepackage{listings}
\usepackage{fancyvrb}
\usepackage{fvextra}
\usepackage{capt-of}
\usepackage{subcaption}
\usepackage{needspace}
\usepackage{placeins}
\usepackage{wrapfig}
\usepackage{lineno}

\lstdefinelanguage{Markdown}{
  basicstyle=\scriptsize,
  sensitive=false,
  morecomment=[l]{\#},   
  morecomment=[s]{```}{```},
  morestring=[b]",        
  morestring=[b]', 
}

\definecolor{darkblue}{rgb}{0, 0, 0.5}
\hypersetup{colorlinks=true, citecolor=darkblue, linkcolor=darkblue, urlcolor=darkblue}

\definecolor{lightred}{HTML}{e99090}
\definecolor{MonsterGreen}{HTML}{2E8B57}
\definecolor{authorHuman}{HTML}{8C564B}
\definecolor{authorClaude}{HTML}{D62728}
\definecolor{authorDeepSeek}{HTML}{FF7F0E}
\definecolor{authorGemini}{HTML}{2CA02C}
\definecolor{authorGPT}{HTML}{1F77B4}
\definecolor{authorKimi}{HTML}{9467BD}

\definecolor{darkorange}{HTML}{CC6600}

\definecolor{codexblue}{HTML}{1F5F8B}

\newcommand{\tmark}[1]{\textsuperscript{#1}}

\newcommand{\storyscope}{\textsc{\small StoryScope}}
\newcommand{\storyscopebig}{\textsc{StoryScope}}

\title{StoryScope: Investigating idiosyncrasies in AI fiction}

\author{Jenna Russell\textsuperscript{\faBook}, 
Rishanth Rajendhran\textsuperscript{\faBook},  
Chau Minh Pham\textsuperscript{\faBook}, 
Mohit Iyyer\textsuperscript{\faBook}, 
John Wieting\textsuperscript{\faMicroscope}\\
University of Maryland, College Park\textsuperscript{\faBook}, 
Google DeepMind\textsuperscript{\faMicroscope}\\
\texttt{\{jennarus, rishanth, chau, miyyer\}@umd.edu}, \texttt{jwieting@google.com}
}

\newcommand{\appendixpromptfigure}[3]{
\par\medskip
\FloatBarrier
\needspace{8\baselineskip}
\begin{center}
\begin{minipage}{\linewidth}
\begin{tcolorbox}[
  colback=gray!5!white,
  colframe=purple,
  title={#2},
  left=1mm,
  right=1mm,
  boxsep=1mm
]
\begingroup
\fvset{
  fontsize=\scriptsize,
  obeytabs=true,
  tabsize=1,
  breaklines=true,
  breakanywhere=true,
  commandchars=\\\{\}
}
\VerbatimInput{#1}
\endgroup
\end{tcolorbox}
\end{minipage}
\end{center}
\captionof{figure}{#2}\label{#3}
\FloatBarrier
\par\medskip
}

\begin{document}

\ifcolmsubmission
\linenumbers
\fi

\maketitle

\begin{abstract}
As AI-generated fiction becomes increasingly prevalent, 
 questions of authorship and originality are becoming central to how written work is evaluated. While most existing work in this space focuses on identifying surface-level signatures of AI writing (e.g., word choice, syntactic structure), we ask instead whether AI-generated stories can be distinguished from human ones \emph{without relying on stylistic signals}, focusing on discourse-level narrative choices such as character agency and chronological discontinuity.
We propose \storyscope, a pipeline that automatically induces a fine-grained, interpretable feature space of
discourse-level narrative features across 10 dimensions (e.g., plot, agents, temporal structure). 
We apply \storyscope\ to a parallel corpus of 10,272 writing prompts, each written by a human author and five LLMs (Claude, DeepSeek, Gemini \nocite{geminiteam2023gemini}, GPT, and Kimi), yielding 61,608 stories, each \textasciitilde5,000 words, and 304 extracted features per story.Narrative features alone achieve 93.2\% macro-F1 for human vs. AI detection and 68.4\% macro-F1 for six-way authorship attribution, retaining over 97\% of the performance of models that include stylistic cues. A compact set of 30 \emph{core} narrative features captures much of this signal: AI stories over-explain themes and favor tidy, single-track plots while human stories frame protagonists' choices as more morally ambiguous and have increased temporal complexity (e.g., flashbacks, nonlinear structure). Per-model \emph{fingerprint} features enable six-way attribution: for example, Claude produces notably flat event escalation, GPT likes using gossip as a plot mechanism, and Gemini defaults to external character description. We find that AI-generated stories cluster in a shared region of narrative space, while human-authored stories exhibit greater diversity. More broadly, these results suggest that differences in underlying narrative construction, not just writing style, can be used to separate human-written original works from AI-generated fiction. We release the \storyscope\ code, 10,272 writing prompts, and 51,336 AI-generated narratives to support future work on narrative analysis and AI authorship.\footnote{Code and data:  \url{https://github.com/jenna-russell/storyscope}}

\end{abstract}
\section{Introduction}
\label{sec:intro}

AI fiction is already under our noses. In March 2026, Hachette, a major publishing house, pulled the horror novel \textit{Shy Girl} after it was flagged as $\sim$78\% AI-generated, the first commercially published novel  canceled over AI allegations. Nearly 20\% of a sample of 14,000 self-published Amazon novels were flagged by Pangram \citep{emi2024technicalreportpangramaigenerated} as largely AI-generated, a figure that jumped 41\% year-over-year.\footnote{\url{https://www.nytimes.com/2026/03/19/books/ai-fiction-shy-girl.html}} Overall, readers are increasingly being misled into purchasing AI-generated books attributed to human authors. If authors are unwilling to self-disclose AI usage, how can we address this issue?

At first glance, this appears to be a detection problem: can we determine whether a given story was written by human or machine?
Existing AI detectors \citep{hans2024spotting, adam2026gptzerorobustdetectionllmgenerated, thai2026editlens} primarily rely on stylistic signals such as word choice and sentence structure, and for good reason: these cues are highly discriminatory. AI-generated text systematically overuses em-dashes, words like ``delve'' and ``tapestry,'' and other surface-level patterns that even simple classifiers detect reliably \citep{sun2025idiosyncrasies, shaib2026measuringaisloptext}.
That said, AI style is increasingly fleeting: GPT~5.4 significantly reduced em-dash usage,\footnote{\url{https://x.com/sama/status/1989193813043069219?s=20}} and fine-tuning to mimic human style drops AI detection rates on creative writing from 97\% to 3\% \citep{chakrabarty2026readerspreferoutputsai}.
Discourse-level narrative features (e.g., plot structure, character agency, information revelation), which we refer to simply as \emph{narrative features} throughout, are far harder to ``humanize,'' as changing them requires significant structural rewrites rather than simple post-hoc edits \citep{namuduri2025qudsim}. 

\begin{figure}[!t]
    \centering
    \includegraphics[width=0.98\textwidth]{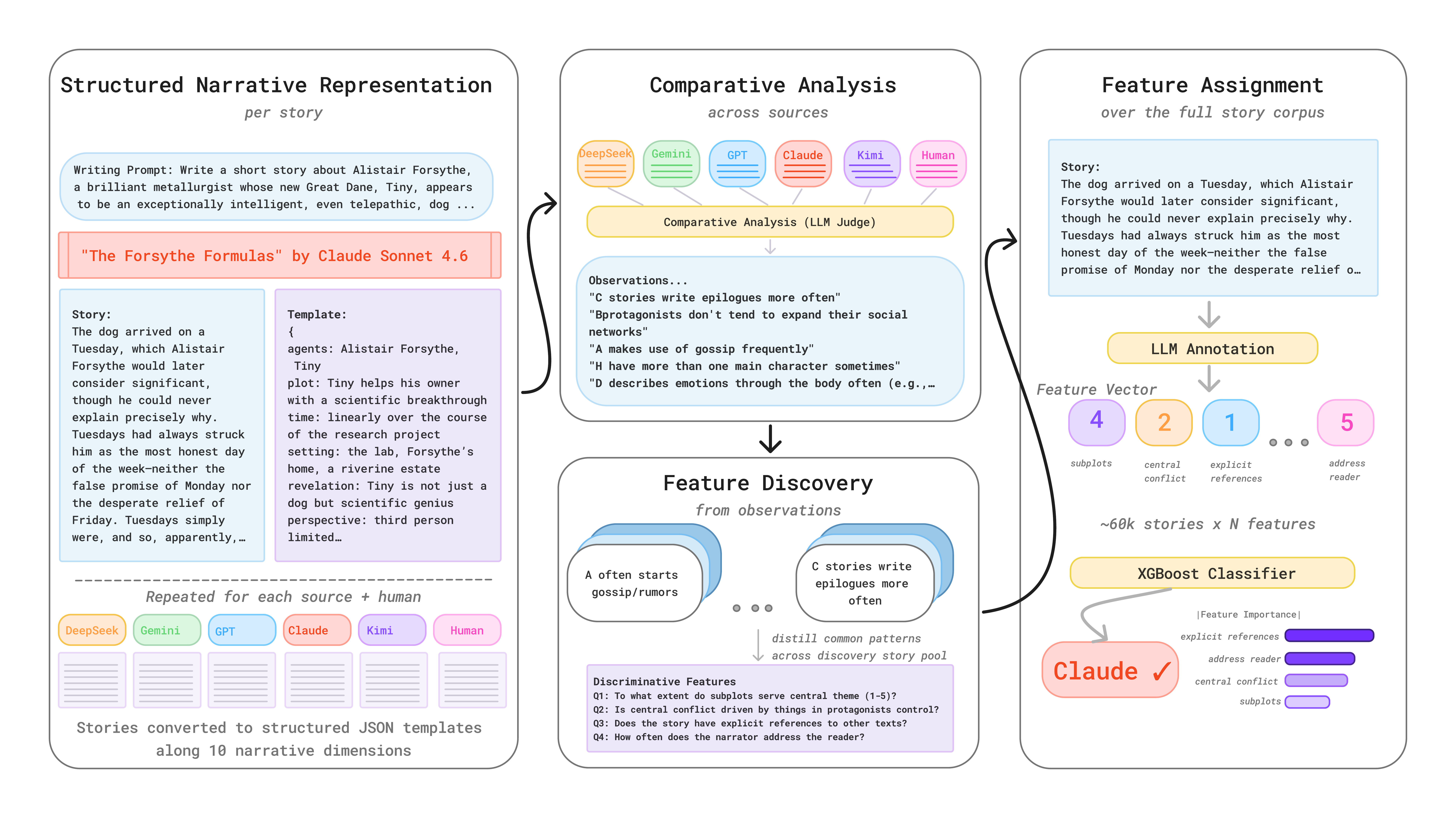}
    \caption{
    Overview of the \storyscope\ pipeline. Stories are converted into structured templates, then compared across sources writing to the same prompt to induce discriminative narrative features, and finally featurized across the full corpus for downstream detection and authorship experiments. Story inspired by "Tiny and the Monster" \citep{sturgeon1947tiny}.}
    \label{fig:pipeline}
\end{figure}

As AI seeps into the writing industry, the question of \emph{what constitutes original work} shifts from how a story is written to how it is conceived. Settled U.S. legal precedent requires that protected works show a minimal degree of originality~\citep{feist1991}; recent guidance from the U.S. Copyright Office clarifies that eligibility depends on sufficient \emph{human} creative control \citep{usco_ai_guidance}. To measure this, we use statistical rarity in a feature space of narrative decisions as a proxy for originality, where less common combinations reflect the broader notion of originality invoked by \cite{torrance1966ttct} and copyright law. We hypothesize that humans and AI models make systematically distinct narrative choices, and that these differences persist even when stylistic cues are removed.

To test this, we introduce \storyscope,  a pipeline that automatically induces
interpretable narrative features grounded in the NarraBench taxonomy \citep{hamilton2025narrabench}, spanning 10 dimensions of narrative structure (character, plot, setting, temporal structure, etc.). 
Applied to a parallel corpus of 10,272 prompts each written by a human author and five LLMs, \storyscope\ yields 304 features and 61,608 featurized stories on which we train classifiers. Our stories average roughly 5,000 words depending on source, enabling extraction of fine-grained narrative features that shorter texts cannot support.

The narrative features alone, withholding all stylistic features (e.g., sentence-level rhythm, figurative language density), achieve a macro-F1 of 93.2\% for the binary human vs. AI detection task, retaining 97\% of the performance of a model with both narrative and stylistic features. In the 6-way authorship attribution task (identifying which source wrote a story), narrative features get a macro-F1 of 68.4\% without style, compared to 77.3\% with style.  When we represent each story as a vector of narrative features, the five AI models occupy a tight cluster that is well-separated from human stories, showing that AI models have converged on a shared narrative space that is systematically separated from human storytelling, and that these changes remain after editing stories for style (macro-F1 93.9\%). Human stories are, on average, rarer in narrative feature space (mean rarity percentile 0.71 vs.\ 0.49 for AI). Each model also exhibits a unique narrative ``fingerprint'': a set of features on which it diverges from the other AI sources and enables fine-grained attribution. 

Our results suggest that the narrative choices underlying AI-generated fiction are distinguishable from those of human authors, even when surface style is removed. Because these features reflect structural decisions rather than lexical ones, they may prove more durable as models continue to evolve. We release the \storyscope\ code, 10,272 writing prompts, 51,336 AI-generated narratives, and narrative features for each story, generated at the cost of \$4.4k USD, to support future work on AI authorship and narrative analysis.\footnote{We do not release human-written stories due to copyright concerns.}

\section{The \storyscopebig\ pipeline}
\label{sec:method}

In this section, we first describe the creation of a dataset that enables comparative analysis of narrative choices, before detailing our pipeline's operation and post-hoc interpretation. 
 
\paragraph{Data} 
Our dataset consists of human-written short stories paired with multiple AI-generated mirrors with similar plots and characters. We extract 10,272 human-written stories from Books3 \citep{presser}.\footnote{We use Books3 \emph{strictly} for academic purposes to understand the narrative differences in recent human-written and AI-generated text and inform discussions on AI-detection, authorship, and copyright policy. Dataset statistics found in \autoref{tab:books3_stats}.} To generate mirrored AI stories, we reverse-engineer writing prompts from each human story by prompting Gemini 2.5 Flash \citep{gemini25flash_modelcard_2025} to infer the underlying premise \citep{li2024selfalignment}.\footnote{Full prompt-extraction template in \autoref{prompt:prompt_generation}. Gemini 2.5 used during writing prompt construction in June 2025. We identify and extract stories from short story anthologies identified via running Gemini 2.5 Flash over titles and blurbs in Books3.} We generate stories from 5 LLMs: Gemini~3 Flash \citep{gemini3flash_model_card_2025}, Kimi~K2.5 \citep{kimiteam2026kimik25visualagentic}, DeepSeek~V3.2 \citep{deepseekai2025deepseekv32pushingfrontieropen}, Claude~Sonnet~4.6 \citep{claudesonnet4.6} and GPT~5.4 \citep{gpt54_system_card_2026}. Together with the human-authored story, this yields six \emph{sources},
 a term we use throughout to refer to each origin uniformly.\footnote{All source identities are anonymized in LLM-facing prompts (see \S\ref{app_sec:featurization}.)}

\subsection{\storyscopebig}
\label{sec:pipeline}
Manually annotating the narrative decisions made within these 60K+ stories (averaging 4,753 words) is infeasible at scale.\footnote{Across the 61,608-story corpus, mean length is 4,753 words. Models refused to generate 24 stories.}
Instead, we leverage LLMs to perform structured analysis over our corpus of stories to create interpretable narrative features. The \storyscope\ pipeline consists of three stages: (1) structured intermediate representations, (2) cross-source comparison, and (3) discourse-level feature discovery, guided by the narrative taxonomy introduced in NarraBench \citep{hamilton2025narrabench}.
Templates convert prose into structured narrative fields, forcing downstream stages to reason over narrative content rather than stylistic surface. The comparative analysis stage distills 600 stories' worth of structured representations into cross-source observations (compressing ${\sim}2.7$M tokens of raw text into ${\sim}686$K tokens) so that feature discovery can generalize from already-identified patterns rather than simultaneously reading, comparing, and formalizing.

\paragraph{Theoretical grounding.}
We transform each story into a template-based representation that preserves narrative elements while abstracting away surface wording, grounding this representation in NarraBench \citep{hamilton2025narrabench}, which defines a taxonomy of narrative dimensions rooted in literary theory. We adopt ten of its twelve aspects: \emph{Agent, Social Network, Event, Plot, Structure, Setting, Time, Revelation, Perspective}, and \emph{Style}.\footnote{We exclude \emph{Paratext} and \emph{Motivation}, which depends on external context unavailable in our setting.}
Each dimension consolidates an established line of narratological work (see Table 5 of \citealp{hamilton2025narrabench}). \storyscope\ searches \emph{within} those dimensions for the specific decisions that separate sources. For example, its temporal dimension reflects long-standing narratological distinctions around narrative order, duration, and frequency from  \citet{genette1980narrative}, while plot arc is heavily inspired by \citet{tian2024humanlevelnarratives}.

\paragraph{Structured narrative representations.}
For each story, we prompt GPT-5.1 \citep{openai2025gpt51_system_card} to extract a structured template organized along these ten dimensions. The extraction uses a zero-shot prompt with a detailed JSON schema specifying the expected fields for each dimension (e.g., character names, roles, and motivations for Agent; causal chains and key events for Event).\footnote{Templating prompt depicted in \autoref{prompt:template_extraction}.} 
The templates serve as a controlled intermediate representation \citep{pham2024topicgpt, wang2023goal}: by converting prose into structured fields, the subsequent comparison stage reasons over narrative content rather than stylistic surface (see \S\ref{app_sec:featurization} for details). 

\paragraph{Comparative story analysis across sources.}
To identify systematic narrative differences, we perform pairwise comparisons across sources writing from the same prompt. We construct a discovery pool
of 600 stories (and their corresponding structured representations) over 100 parallel randomly selected writing prompts. The discovery pool is held out from the main corpus and kept small because this stage uses GPT-5.1 with high reasoning effort over long inputs; 600 stories balance narrative diversity against cost. For each prompt, we present all six templates to GPT-5.1 and ask it to produce a structured comparative analysis: a structured JSON containing per-source dimension notes (e.g., how each source handles character motivation), cross-source comparisons highlighting where sources diverge, and an executive summary of recurring patterns.\footnote{We batch multiple prompts per analysis (mean 3.1), details in \S\ref{app_sec:featurization}.} The comparative analyses serve as the raw material from which we derive our feature taxonomy.

\begin{table*}[!t]
\centering
\scriptsize
\begin{tabular}{@{}r p{0.18\textwidth} p{0.25\textwidth} l l p{0.23\textwidth}@{}}
\toprule
\textbf{\#} & \textbf{Feature} & \textbf{Question} & \textbf{Dim.} & \textbf{Type} & \textbf{Response Options} \\
\midrule
\multicolumn{6}{l}{\textit{AI}} \\
1 & Thematic Explicitness and Moralizing & How explicitly does the story articulate its themes or morals? & SIT & scale & 1--5 \\
2 & Agency in Resolution $\rightarrow$ protagonist choice & Is resolution driven by protagonist's choices or external events? & PLT & cat & protagonist\_choice, mixed, external\_fate \\
3 & Narratorial Thematic Commentary $\rightarrow$ yes & Does the narrator explicitly comment on themes beyond characters' perspectives? & SIT & binary & no, yes \\
\midrule
\multicolumn{6}{l}{\textit{Human}} \\
1 & Intertextual Strategy Types $\rightarrow$ explicit named reference & What kinds of intertextual engagement does the story employ? & SIT & multi & explicit named, retelling, pastiche, myth/religion, self referential \\
2 & Depth of Recontextualization After Surprise & How extensively does a revelation force reinterpretation of earlier scenes? & REV & scale & 1 (none)--5 (complete re-reading) \\
3 & Degree of Chronological Discontinuity & How often does the narrative jump across time? & TMP & scale & 1--5 \\
\bottomrule
\end{tabular}
\caption{Examples of core features, see AI core features in \autoref{tab:core_ai_features}, human in \autoref{tab:core_human_features}, and all 30 with human/AI means in \autoref{tab:core_features_themed}.
NarraBench dimensions include situatedness (SIT), plot (PLT), revelation (REV) and temporal structure (TMP).}
\label{tab:core_feature_excerpt}
\vspace{-8pt}
\end{table*}

\paragraph{Feature discovery.}
We extract interpretable narrative features from the comparative analyses by prompting GPT-5.1 to propose discriminative features within each NarraBench dimension.\footnote{We use 10 specialized expert prompts, one per NarraBench dimension (see \S\ref{app_sec:application}).} Each expert prompt is grounded in a specific NarraBench aspect and instructs the model to propose features as closed-form questions with discrete answer choices. We constrain features to five response types (categorical, ordinal, scale, binary, and multi-select) to support interpretable downstream modeling. 
To improve coverage, we run the discovery process three times and take the union of all proposed features, yielding 408 candidates. We then deduplicate via embedding-based clustering, retaining the feature nearest each cluster centroid, resulting in $d = 304$ features after merging 65 clusters.\footnote{Each feature is encoded with F2LLM-4B \citep{zhang2025f2llm} and clustered at cosine similarity threshold 0.85, details in \S\ref{app_sec:featurization}.} Each feature is defined as a specific axis of narrative variation (e.g., the degree to which character motivation is stated explicitly versus left implicit, or the extent to which a story's timeline departs from chronological order). The final taxonomy spans five response types: categorical (124), ordinal (59), scale (45), binary (44), and multi-select (32).

\subsection{\storyscope\ Interpretability}

\paragraph{Feature assignment.}
We apply the features to the full dataset of 61{,}608 stories across all 10{,}272 prompts. For each story, we present the full text along with the relevant feature definitions to Gemini~3 Flash (with minimal thinking) and obtain a value for each of the 304 features,\footnote{Gemini-3 has high repeatability over 5 independent runs (Krippendorff's $\alpha = 0.90$), human-validation on a 240 feature subset has a mean human--model Cohen's $\kappa = 0.84$ (details in \S\ref{app_sec:human_validation}).} yielding a raw narrative-assignment vector $\mathbf{z} = (z_1,\dots,z_d)$ with $d=304$ semantic features for every story in the corpus. 
The classifier does not consume $\mathbf{z}$ directly: after 
encoding each feature according to its type
(one-hot for categorical, multi-hot for multi-select, numeric for ordinal and scale, binary as-is), 
we obtain an encoded input vector $\mathbf{x} \in \mathbb{R}^{D}$, where $D$ depends on the feature subset and the expansion induced by categorical and multi-select features.

\paragraph{Defining core and fingerprint features.}
Not all features contribute equally to distinguishing sources. To identify which features are robust human-AI markers versus source-specific cues, we train XGBoost classifiers \citep{chen2016xgboost}.
Gradient-boosted trees pair naturally with SHAP \citep{lundberg2017shap} to yield exact, per-feature importance decompositions, letting us trace every prediction back to specific narrative decisions.
We assess importance and stability through bootstrap SHAP analysis ($B{=}50$ iterations with prompt-level resampling) and assign each feature one of three roles:\footnote{This procedure yields 30 core features and 75 fingerprint features. Full selection criteria, equations, and numerical thresholds are detailed in \S\ref{app_sec:feature_roles}.} (1) \textbf{Core Features:} stable and important in the \emph{binary} human-vs-AI task, with a strong, consistent separation that holds across all five AI models (examples in \autoref{tab:core_feature_excerpt}). Each is signed as human-leaning or AI-leaning based on the direction of the gap; (2) \textbf{Fingerprint features:} identified from the 6-way \emph{multiclass} task. A feature qualifies when its SHAP importance is concentrated in a single source class, and that source's observed values visibly differ from others; (3) Features satisfying neither set of criteria are excluded from role-based analyses.
\section{Experiments}
\label{sec:experiments}

\paragraph{Classification.}
We train XGBoost classifiers on encoded feature vectors $\mathbf{x}$ over 52,707 stories (8,788 prompts) for two tasks: (1) binary detection (human vs.\ AI) and (2) 6-way authorship attribution (human, Claude, GPT, DeepSeek, Gemini, Kimi), evaluated on a held-out test set of 1,384 prompts (8,301 stories) with macro-F1 and AUPRC as the primary binary metrics.\footnote{Hyperparameters are selected via grid search on the validation split, \textit{Binary}: $n_\text{est}{=}420$, depth${=}8$, $\lambda{=}2.0$, and 5:1 positive class weight; 6\textit{-way}: $n_\text{est}{=}500$, depth${=}7$, $\lambda{=}1.0$, evaluation details in \S\ref{app_sec:experiments}.} Nominal features are one-hot encoded, multi-select features are multi-hot encoded, and ordinal / scale features retain numeric encoding. 
For the full Narrative+Style this yields $D=1108$ encoded columns; the Narrative variant uses $D=958$, and Style Only uses $D=129$. All evaluations use prompt-level grouping to prevent train/test leakage.

\paragraph{Feature variants.}
A central question of this work is whether narrative \emph{choices} carry discriminative signal independent of surface style. To test this, we evaluate the following: \textbf{Narrative} (257 features) includes all narrative features across nine NarraBench dimensions, excluding all 39 features in the style dimension plus 8 features from other dimensions flagged as style-related, 47 in total.\footnote{We flagged features using an LLM audit of a feature's reliance on stylistic cues, details in \S\ref{app_sec:featurization}.} This is our primary model; it tests whether narrative structure alone suffices for the detection and attribution tasks. \textbf{Style Only} (39 features) uses only the style features. \textbf{Narrative + Style} (304 features) uses all features across all 10 NarraBench dimensions. \textbf{Core Only} 
(30 features) uses the subset of narrative features identified as universally important and stable (\S\ref{app_sec:feature_roles}). \textbf{Core + Fingerprint} (101 features) adds source-specific fingerprint features to the core set.

\paragraph{Baselines.}
We compare against four text-based baselines that operate on raw story text rather than extracted features: (1)~ModernBERT \citep{warner2025modernbert}, a transformer fine-tuned on our training set (max length 512, 3 epochs); (2)~Stylometric+XGB, XGBoost on 144 hand-crafted stylometric features (character n-grams, POS distributions, readability scores); (3)~TF-IDF+XGB, XGBoost on 5{,}000 TF-IDF features; and (4)~Binoculars \citep{hans2024spotting}, a zero-shot AI-text detector. These baselines establish an upper bound on raw-text detection; we are interested in understanding how closely our Narrative model performs.

\section{Detecting AI from narrative features}
\label{sec:detectability}

We evaluate whether narrative features can separate human-written from AI-generated fiction.
Narrative features alone achieve a macro-F1 of 93.2\%, just 2.8 points below the model with both style and narrative features in macro-F1 (95\% prompt-bootstrap CI 2.09--3.54). A set of 30 \emph{core} features retains about 91\% of the narrative model's macro-F1 (84.8\%). We ask if \storyscope\ can still detect AI narratives after stylistics alterations, finding it robust to edits. 

\begin{wraptable}{r}{0.48\columnwidth}
\vspace{-8pt}
\centering
\small
\begin{tabular}{@{}lccc@{}}
\toprule
\textbf{Method} & \textbf{Size} & \textbf{F1} & \textbf{AP} \\
\midrule
\multicolumn{4}{l}{\textit{Narrative features (ours)}} \\
\quad Narrative & 257 & 93.2 & .959 \\
\quad Core Only  & 30  & 84.8 & .828 \\
\quad Core+FP    & 101 & 91.1 & .934 \\
\midrule
\multicolumn{4}{l}{\textit{Style ablations}} \\
\quad Narr.\ + Style & 304 & 96.0 & .982 \\
\quad Style Only     & 39  & 85.8 & .867 \\
\midrule
\multicolumn{4}{l}{\textit{Text-based baselines}} \\
\quad ModernBERT      & ---   & 99.9 & 1.00 \\
\quad Stylometric+XGB & 144   & 99.8 & .999 \\
\quad TF-IDF+XGB      & 5,000 & 99.7 & .999 \\
\quad Binoculars      & ---   & 55.9 & .404 \\
\midrule
\multicolumn{4}{l}{\textit{Edited Stories}} \\
\quad LAMP      & ---   & 93.9 & .988 \\
\bottomrule
\end{tabular}
\caption{Binary human vs. AI classification results. F1 = macro-F1 (\%); AP = AUPRC.}
\label{tab:binary_results}
\vspace{-8pt}
\end{wraptable}

\paragraph{Style signals are stronger but narrative features close the gap.}
Models built using only the style features achieve comparable performance to the core-only model (85.8\% and 84.8\% macro-F1 respectively), confirming that style remains a strong detection cue, especially in long-context settings. Most automatic AI detection models have higher performance on longer text \citep{bao2024fastdetectgpt, xu2025trainingfree}. The narrative model is unchanged with length: on a human-length-matched test subset, it scores 93.2\% macro-F1 both before and after matching (see \S\ref{app_sec:length_confound}). While the strongest supervised text-based baselines achieve near-perfect separation ($\geq$99.7\% macro-F1, with ModernBERT at 99.9\%), zero-shot Binoculars is much weaker (55.9\% macro-F1). Our narrative+style model nearly matches the strongest baselines (96.0\% macro-F1), with the added benefit of full interpretability.\footnote{We also ask if model style is affected by memorization of famous stories, finding that after filtering out likely memorized stories performance remains the same (see \S\ref{app_sec:memorization}).} Narrative features alone recover 97\% of the combined model's macro-F1.

\paragraph{Core features retain most of performance.}
Based on just 30 core features, our core model retains substantial human vs. AI separation, reaching 84.8\% macro-F1 and 0.828 AUPRC. Representative core-feature definitions were introduced earlier in \autoref{tab:core_feature_excerpt}; see \S\ref{app_sec:feature_catalogs} for the complete lists. Adding fingerprint features raises performance to 91.1\% macro-F1 and 0.934 AUPRC, recovering much of the gap to the full narrative model while using less than half of the features. Our results suggest that the human-AI boundary lives in a compact subset of narrative decisions, while fingerprint features contribute model-specific separability. 

\begin{figure}[t]
\centering
\includegraphics[width=\textwidth]{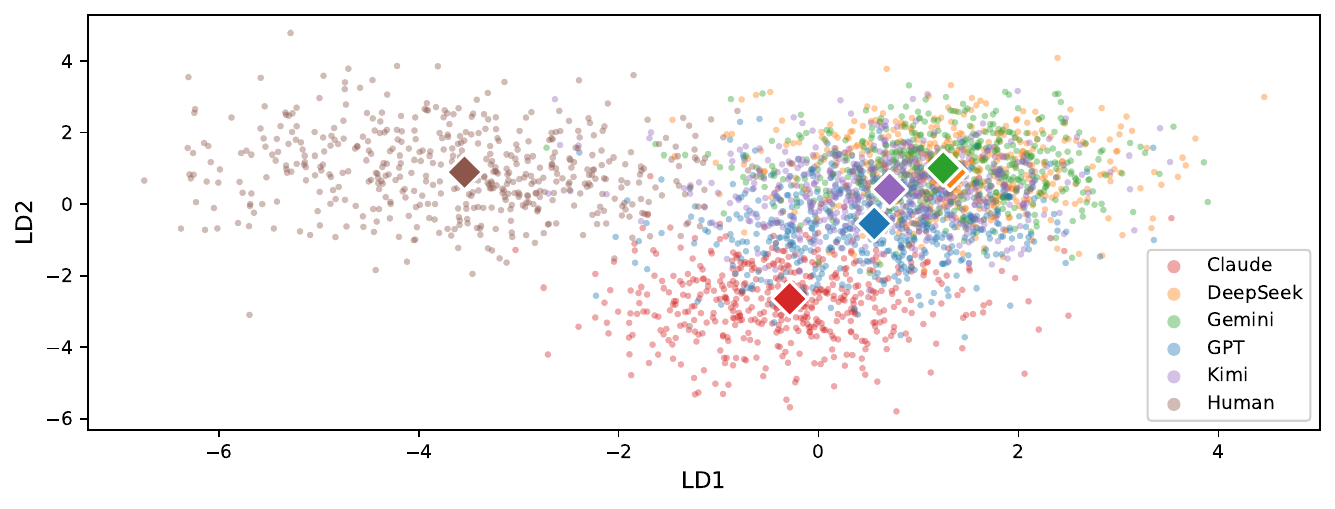}
\caption{Projection of narrative feature vectors onto the first two linear discriminant components. Human writing occupies a distinct region; the five AI models cluster together. Claude is the most distinct of the 5 AI models, Gemini and DeepSeek the nearest neighbors.}
\label{fig:lda_6way}
\vspace{-8pt}
\end{figure}

\subsection{What core narrative choices separate human from AI writing?}

\paragraph{AI over-explains its themes.} 
AI stories are more explicit and moralizing, (roughly 20\% higher on 1-5 scale), with tighter thematic unity and more central moral questions. Narrators explicitly explain the story's theme 77\% of the time, versus 52\% for humans: a grieving character's arc will typically end with the narrator stating the lesson learned.
AI dialogue serves philosophical debate more often (59\% vs.\ 34\%), and references to other works tend to be vague allusions (72\% vs.\ 50\%) rather than specific, named references. The pattern is one of over-determination: AI spells out meaning rather than trusting the reader to infer it.

\paragraph{Human authors subvert linearity.} 
AI stories exhibit tighter causal chains, more protagonist-driven resolutions (69\% vs.\ 46\%), and far fewer subplots (79\% ``no subplots'' vs.\ 57\%). AI resolutions favor internal understanding or acceptance (47\% vs.\ 27\%), whereas humans are more comfortable with ambiguous endings. Humans use more time jumps, flashbacks and flash-forwards, and nonlinear structure to delay key revelations. AI favors single-track narratives with fewer loose ends; human stories are messier, with time jumps and disjointed causal chains---a human mystery might open at the funeral and spiral backward through decades, while AI tells the same story from first clue to the grand reveal.

\paragraph{AI over-writes the body and senses.} AI overwhelmingly conveys emotion through physical sensations and bodily metaphors (81\% vs.\ 38\% human), deploys more smell-based imagery (82\% vs.\ 57\%), and uses setting as a reflection of characters' inner states more heavily. AI pays closer attention to physical environment and characters' inner mental states, showing emotion through bodies and environments rather than naming feelings directly. Where a human author might write that a character ``felt afraid,'' AI renders fear as a tightening chest, cold sweat, and dimming lamplight. Humans use explicit emotion labels 29\% of the time versus just 8\% for AI.

\paragraph{Human authors engage the outside world.} 
Humans reference specific texts and authors at nearly double the AI rate (47\% vs.\ 24\%) and balance explicit with implicit references more evenly (37\% ``balanced mix'' vs.\ 16\%), whereas AI generally sticks to vague allusions and avoids naming real brands, places, or works. Humans break the fourth wall far more often (67\% vs.\ 39\%) and address the reader directly more frequently (28\% vs.\ 7\%). Human writing acknowledges its audience as a co-participant (e.g., an aside to ``you, dear reader''); AI writes as though no one is watching.

\paragraph{AI writing has less diverse narrative features.}
Human stories draw from a broader narrative repertoire: they span more locations, carry more dialogue relative to narration, integrate more subplots into overarching themes (42\% vs.\ 21\%), and present morally ambivalent protagonists more often (59\% vs.\ 38\%), resisting the pull to a narrow set of defaults.

\subsection{Do narrative features still work after stylistic changes?}

Style is the most obvious surface-level cue distinguishing AI generations from human writing, and style-based detectors are known to be brittle to paraphrasing and light editing~\citep{saakyan2026death}. We ask whether narrative features are robust to such edits: if an AI-generated story is stylistically edited to remove surface artifacts, does our narrative classifier still detect it?

We test this using \citet{chakrabarty2024salvaged}'s span-level rewriting framework (LAMP), which identifies and rewrites seven categories of AI writing artifacts (e.g., clich\'{e}, redundant exposition, purple prose) using 25 few-shot examples from professional writers. We apply this to 278 Gemini-generated stories, using Gemini itself as the rewriter, and evaluate our narrative classifier on these edited stories.

\paragraph{Surface editing barely affects narrative detection.} After span-level artifact removal, the narrative model detects edited Gemini stories at 93.9\% macro-F1 (AUPRC 0.988), 
compared to 95.5\% macro-F1 (AUPRC 0.996) 
on the original unedited stories, a drop of only 1.6 points. This near-zero effect indicates that narrative features are largely orthogonal to the surface prose artifacts that LAMP targets: editing out clich\'{e}d phrasing or purple prose does not alter the structural narrative choices (causal linearity, thematic explicitness, sensory over-description) that drive our classifier.

\section{Pinpointing each source's writing style}
\label{sec:authorship}

The binary task asks whether a story is human or AI; we now ask if we can use narrative features 
to detect exactly \emph{which} source wrote a story. Using the same features and classifier, we train a 6-way model over all six sources. Our narrative model reaches 68.4\% macro-F1, indicating that narrative structure alone still carries weighty authorship signal, though the AI-authorship boundaries overlap much more than the binary human-AI boundary.

\begin{wraptable}{r}{0.48\columnwidth}
\vspace{-8pt}
\centering
\small
\begin{tabular}{@{}lccc@{}}
\toprule
\textbf{Method} & \textbf{Size} & \textbf{F1} & \textbf{Acc.} \\
\midrule
\multicolumn{4}{l}{\textit{Narrative features (ours)}} \\
\quad Narrative          & 257   & 68.4 & 68.4\% \\
\quad Core Only          & 30    & 46.5 & 46.8\% \\
\quad Core+FP            & 101   & 63.4 & 63.6\% \\
\midrule
\multicolumn{4}{l}{\textit{Style ablations}} \\
\quad Narr.\ + Style     & 304   & 77.3 & 77.3\% \\
\quad Style Only         & 39    & 60.4 & 60.5\% \\
\midrule
\multicolumn{4}{l}{\textit{Text-based baselines}} \\
\quad ModernBERT         & ---   & 99.8 & 99.8\% \\
\quad Stylometric+XGB    & 144   & 99.6 & 99.6\% \\
\quad TF-IDF+XGB         & 5,000 & 99.5 & 99.5\% \\
\bottomrule
\end{tabular}
\caption{6-way authorship attribution on the test set. F1 = macro-F1 (\%). As AI models converge, the attribution task proves much more difficult than the binary detection task.}
\label{tab:multiclass_results}
\vspace{-20pt}
\end{wraptable}

\paragraph{Narrative features retain substantial attribution signal.} 
\autoref{tab:multiclass_results} shows macro-F1 and accuracy results for 6-way attribution. Our narrative model achieves 68.4\% macro-F1, well above the 16.7\% chance baseline but below the narrative+style model at 77.3\%. Narrative exceeds Style Only by 8.0 macro-F1 points (95\% prompt-bootstrap CI 6.7--9.2), while Core+Fingerprint exceeds Core Only by 16.8 points (15.7--17.9), consistent with style and fingerprint features adding complementary attribution signal. Text-based baselines remain far stronger on raw text, all reporting macro-F1s of at least 99.5\%.

\paragraph{Claude and GPT most distinctive of AI models.}
\autoref{tab:per_class_f1} breaks down per-class F1 for the 6-way task. Human is the most distinctive source (93.0\% F1 with style, 88.5\% without), followed by Claude (89.3\% / 77.1\%) and GPT (82.1\% / 73.0\%). DeepSeek, Gemini, and Kimi form a more confused cluster (65.8--66.8\% F1 with style). The gap widens in the narrative-only model: human and Claude retain strong separability while the bottom three models fall to 55.2--59.6\%, suggesting that narrative structure alone is less diagnostic for distinguishing among models that make similar storytelling choices. 

\begin{wrapfigure}{l}{0.5\columnwidth}
\vspace{-12pt}
\centering
\includegraphics[width=0.48\columnwidth]{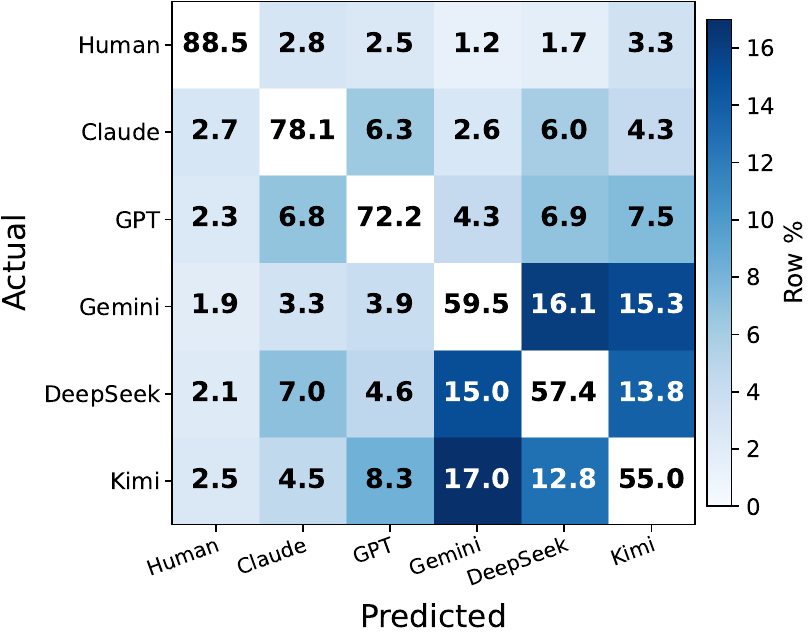}
\caption{Confusion matrix for authorship attribution (narrative model) as a percentage (\%). Misclassifications concentrate among AI models, particularly DeepSeek--Gemini--Kimi.}
\label{fig:confusion_matrix}
\vspace{-35pt}
\end{wrapfigure}

\paragraph{AI convergence separates from human narratives.} 
The five AI sources occupy overlapping regions of narrative feature space, separate from human authors (\autoref{fig:lda_6way}). 
The six most-confused source pairs in our narrative-only 6-way classifier are exclusively AI$\leftrightarrow$AI (\autoref{fig:confusion_matrix}); the largest pair (gemini$\leftrightarrow$deepseek, 222 and 207 stories) dwarfs the most common human misclassification (human$\rightarrow$kimi, 46).
Working in the
z-scored encoded feature space with Euclidean distance, mean human-AI centroid distance is 1.6$\times$ the mean AI-AI centroid distance (6.6 vs.\ 4.3). Even the closest human-AI centroid pair is farther apart than the most distant AI-AI pair (6.2 vs.\ 6.0), indicating that human stories occupy a distinct region rather than merely a broader version of the AI cluster. Human stories are also more dispersed: their mean distance to the human centroid is 22\% greater than the average AI radius (33.2 vs.\ 27.4) and their median 10-nearest-neighbor radius is 1.13$\times$ larger (33.1 vs.\ 29.2).

\paragraph{Human narratives are rarer.}
We measure per-story rarity as the mean Euclidean distance to a story's 25 nearest neighbors. Human stories have a higher mean rarity percentile than AI stories (0.71 vs.\ 0.49; Cohen's $d$\,=\,0.83 indicating a substantial difference), and humans are overrepresented in the rarest tail: 24.7\% of all human 
  stories fall in the top 10\% rarest stories corpus-wide, compared to just 7.1\% of AI stories (top 1\%: 3.0\% human vs.\ 0.6\% AI). At the prompt level, the human story is ranked the rarest of all six story versions 57.8\% of the time (vs.\ 16.7\% by chance) (per-source distributions in \S\ref{app_sec:rarity_analysis}).

\paragraph{Claude keeps it cool.}
Claude has the most distinctive narrative profile of the five LLMs. Its stories are defined by its restraint: event intensity escalates less than in any other source, and narrative voice is the most uniform. Claude takes a reverent/continuist approach to literary tradition, honoring and extending storytelling conventions rather than subverting or challenging them (62\% of Claude stories vs.\ 39--56\% across other sources). It favors epilogues and avoids dream sequences, producing careful, consistent stories that favor quiet endings over 'avalanche' endings.

\paragraph{GPT likes to gossip.}  
GPT centers on socially-oriented storytelling: gossip and rumor as a plot mechanism (64\% vs.\ 44--55\% for other sources), a  tendency to frame stories as reflections on events from years or decades ago, and ensemble-heavy social networks matching human levels. GPT subverts expectations more than other AI (41\% vs.\ 27--36\%) and leaves reconciliations ambiguous. 

\paragraph{Gemini, DeepSeek, and Kimi are `triplets'.} 
Despite forming a more confused cluster, each still has individual quirks. DeepSeek front-loads crucial context that other sources leave until later. Gemini produces the tidiest endings, extended denouements, and the bleakest settings (88\% tagged bleak and oppressive). Kimi has the fewest fingerprints and lowest F1, sitting at the generic center of the AI distribution with no distinctive narrative choices. 

\section{Related Work}
\label{sec:related_works}

\paragraph{AI slop and detection.}
LLM text exhibits systematic lexical and syntactic patterns~\citep{shaib2026measuringaisloptext, kobak2024delving, shaib2025learning}, and parallel-corpus studies show LLMs converge on a narrow grammatical and rhetorical style distinct from human variation~\citep{reinhart2025llmswrite}. Detection methods range from stylometric and linguistic features~\citep{li2025styledecipherrobustexplainabledetection, uchendu2025stylometric, tripto2025chokepoints} to ML classifiers on creative fiction~\citep{mcglinchey2024detecting, najjar2025leveraging, mcgovern2025fingerprints, huang2024authorship}, with expert humans also achieving near-perfect accuracy~\citep{russell2025people}. Detecting partial AI edits to human text remains harder~\citep{thai2026editlens, chen2024imbd, he2024detree}.

\paragraph{Originality and creativity.}
LLM outputs are less novel across n-gram~\citep{padmakumar2026measuring}, psychometric~\citep{chakrabarty2024artifice, koivisto2023humans, gomezrodriguez2024duel}, and creativity-theoretic measures~\citep{boden2004creative, colton2012computational, franceschelli2023creativity}. Narrative theory provides structural grounding for generation and understanding~\citep{liu2026narrativesurvey}: LLMs can extract discourse-level features such as POV, temporality, and focalization~\citep{piper2024narrative}, and prompt-based comparisons reveal that human stories contain more cultural nuance, emotional ambiguity, and unexpected twists than GPT-generated counterparts~\citep{begus2024experimental}. Network analyses reveal simpler social structures in AI fiction~\citep{nonaka2025network}. Fine-tuning on literary corpora produces preferred text~\citep{ chakrabarty2026readerspreferoutputsai} but activates verbatim memorization~\citep{liu2026alignmentwhackamolefinetuning}; constrained generation from human text fragments further blurs the authorship boundary~\citep{pham2025frankentext}. Across benchmarks~\citep{hou2026creativityprism, paech2024eqbench, zhang2025noveltybench, saakyan2026death, wadhwa2026createtestingllmsassociative, nguyen2025divergentconvergentthinkinglargelanguage}, LLMs consistently reduce collective diversity~\citep{doshi2024generative, jiang2025artificial}, with repeated plot elements across generations~\citep{xu2025echoes}.
\section{Conclusion}
\label{sec:conclusion}
We introduce \storyscope, a pipeline for extracting interpretable narrative features at scale, and show that these features alone achieve 93.2\% macro-F1 for human vs. AI detection and 68.4\% for six-way attribution across 61,608 stories, retaining performance over 97\% of models that also include stylistic signals. A compact set of 30 core features captures much of this separation: AI stories are systematically more thematically explicit, causally tidy, and temporally linear, while human stories show greater structural diversity and occupy a rarer, more dispersed region of narrative space. As surface-level signatures become increasingly transient, often removed by newer model versions or simple post-hoc edits, narrative features offer a more durable basis for authorship analysis, since altering them requires significant structural rewrites. Our features provide a measurable proxy for narrative uniqueness that complements existing detection tools.

\section*{Acknowledgments}
\label{sec:acknowledgements}
We thank the University of Maryland Computational Linguistics and Information Processing (CLIP) Lab for their feedback and support. This project was partially supported by awards IIS-2046248 and IIS-2312949
from the National Science Foundation (NSF). We thank Google for a Cloud Credit award and Pangram Labs for an OpenRouter credit award, both of which enabled this research.

\section*{Ethics Statement}
\label{sec:ethics}

\paragraph{Use of Books3.} 
We acknowledge the copyright issues related to the Books3 dataset \citep{presser} and do not endorse its use for model training or commercial text generation.
The use of this dataset in our paper is restricted to academic purposes only and is meant to understand the narrative differences in human-written and AI-generated text to help inform discussions on AI-detection, authorship, and copyright policy.

\paragraph{AI Disclosure.}
Large language models and coding agents (Claude Code and Codex) are used to aid with and polish writing and generate some tables and plots.

\bibliography{colm2026_conference}
\bibliographystyle{colm2026_conference}

\appendix


\section{Data}
\label{app_sec:data}

\begin{figure}[t]
\centering 
\includegraphics[width=\linewidth]{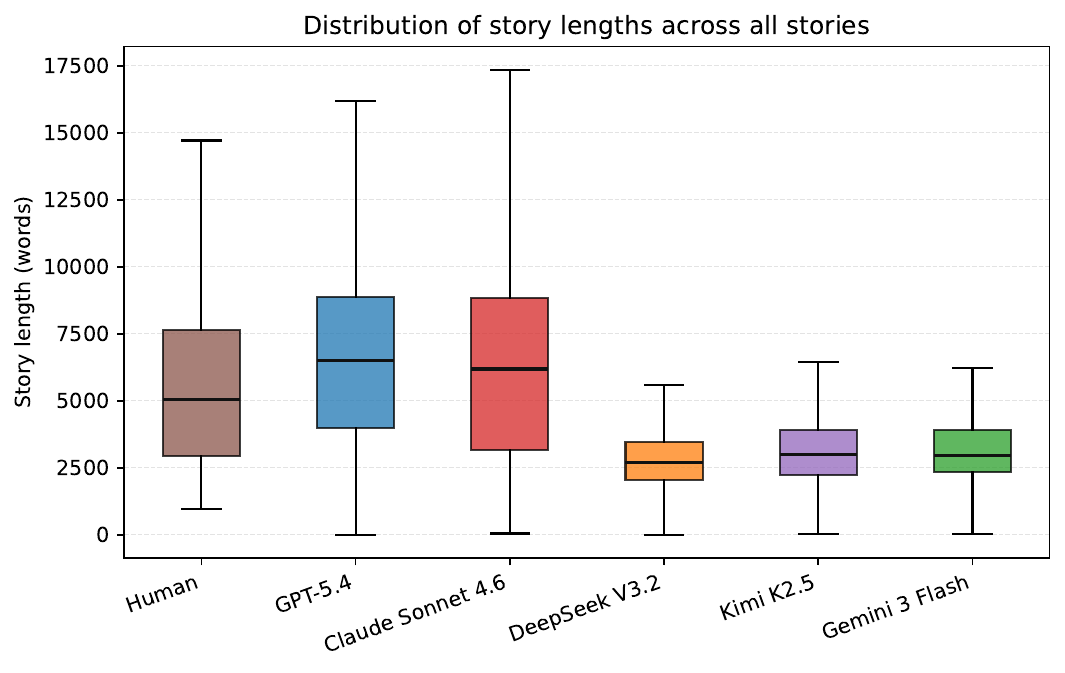}
\caption{Boxplots of story lengths across all stories in the finalized six-sources dataset, shown separately for the human corpus and each model. Boxes show the interquartile range, center lines show medians, and whiskers show the non-outlier range.}
\label{fig:appendix_story_length_boxplots}
\end{figure}

Information about our human-written corpus depicted in \autoref{tab:books3_stats}. Information on average length of generated stories, as well as how well models adhered to length based instructions are in \autoref{tab:ai_stories_stats} and shown in \autoref{fig:appendix_story_length_boxplots}. Story generation costs varied widely by model and was completed for a total cost of roughly \$2800 USD. Extracting features from Gemini 3 Flash \citep{gemini3flash_model_card_2025} over the entire story-corpus cost roughly \$1600 USD. All stories were generated with a maximum length of 128,000 tokens, except for Gemini 3 Flash, whose maximum token limit is 65,536. 

\begin{table}[t]
\centering
\scriptsize
\begin{tabular}{p{0.58\linewidth}r}
\toprule
\textbf{Statistic} & \textbf{Value} \\
\midrule
Prompts / human stories & 10,272 \\
Target words (mean / median) & 6,242 / 5,000 \\
Target words (IQR) & 3,000--7,000 \\
Human story words (mean / median) & 6,403 / 5,035 \\
Human story words (IQR) & 2,930--7,650 \\
\bottomrule
\end{tabular}
\caption{Human-corpus summary. All stories extracted from Books3..}
\label{tab:books3_stats}
\end{table}

\begin{table*}[t]
\centering
\scriptsize
\resizebox{\textwidth}{!}{%
\begin{tabular}{lrrrrrrr}
\toprule
\textbf{Model} & \textbf{$n$} & \textbf{Coverage} & \textbf{Mean words} & \textbf{Median words} & \textbf{Mean $\Delta$ words} & \textbf{Mean abs.\ \% err.} & \textbf{Within 10\%} \\
\midrule
Gemini 3 Flash & 10,261 & 99.9\% & 3,155 & 2,967 & -3,089 & 36.9\% & 19.4\% \\
GPT-5.4 & 10,272 & 100.0\% & 6,651 & 6,515 & 409 & 40.2\% & 9.6\% \\
Claude Sonnet 4.6 & 10,259 & 99.9\% & 6,817 & 6,187 & 575 & 16.6\% & 38.4\% \\
DeepSeek V3.2 & 10,272 & 100.0\% & 2,946 & 2,693 & -3,296 & 39.9\% & 12.8\% \\
Kimi K2.5 & 10,272 & 100.0\% & 3,274 & 3,005 & -2,968 & 35.2\% & 15.6\% \\
\bottomrule
\end{tabular}
}
\caption{Length and coverage statistics for AI-generated stories in the finalized six-model dataset. Word-target adherence is computed against the explicit ``approximately $N$ words'' instruction embedded in each prompt; $\Delta$ is signed model-word-count minus target-word-count.}
\label{tab:ai_stories_stats}
\end{table*}

\section{\storyscopebig\ Details}
\label{app_sec:featurization}

\paragraph{Blinding Protocol}
Throughout the pipeline, sources identities are anonymized in all LLM-facing prompts. Within pairwise template comparisons, presentation order is randomized to reduce positional bias while keeping the comparisons reproducible.
\paragraph{Are templates truly needed?}
We ran early stage testing on the full pipeline with and without template extraction, the only difference being whether the comparative analysis operates on templates or raw story text. 
The two variants discover qualitatively different features: comparing the top-20 discriminative features, only 6 overlap. The direct pipeline's features are style-heavy (humor usage, vocabulary register, allusion types, dominant imagery), while the template pipeline's are structure-heavy (emotional arcs, relationship trajectories, event density, flashback usage).
\paragraph{Meta-Analysis Batching Details}
 The cross-author comparative analysis operates over 100 development prompts, each with six sources templates. We batch prompt sets greedily by estimated token count, treating the combined size of the six templates for a prompt as the batching unit; this yields 32 batches, averaging 3.1 prompts per batch. 
 A parallel Gemini-3-Flash run permits fewer, larger batches but produces substantially fewer features, so we retain the GPT-5.1 outputs. 
\paragraph{Feature deduplication.}
Feature discovery produces 408 candidate features, many of which capture overlapping concepts (e.g., ``metanarrative self-awareness'' and ``self-reflexive narration''). We deduplicate them with embedding-based clustering: each feature is represented by its name, question, and detection method, encoded with F2LLM-4B \cite{zhang2025f2llm}, then clustered with single linkage at cosine threshold 0.85. We keep the feature nearest each cluster centroid as the representative, reducing the taxonomy from 408 to 304 features (25.5\% reduction) and merging 65 multi-feature clusters. 
We use 0.85 because 0.90 left many clear duplicates unmerged (372 features total), whereas thresholds above 0.85 quickly stopped merging semantically overlapping features that differed only slightly in wording.
\paragraph{Feature Taxonomy Details}
The final taxonomy contains $d = 304$ interpretable narrative features spanning five response types
\autoref{tab:feature_taxonomy_types} summarizes the final distribution of feature types and gives one full illustrative feature entry for each type.

\paragraph{Style boundary.}
 Style features concern sentence- and phrase-level texture (diction, syntax, rhythm, figurative density, tonal register, and prose-level allusion), whereas non-style features concern narrative content and structure (events, causality, closure, character roles, relationship topology, setting, and temporal arrangement). For gray-zone cases, our rule is simple: if a feature can only be answered from prose texture, it is style; if it is primarily about narrative content and prose choices are incidental, it is non-style. \emph{Sensory Density}, \emph{Depth of Interior Access}, and \emph{Chronological Discontinuity} fall on the non-style side of this rule. We conducted a style-dependence audit over all 304 taxonomy features, prompting GPT-5.4 \citep{gpt54_system_card_2026} to rate each as \texttt{high}, \texttt{medium}, or \texttt{low} style dependence. For the strict narrative variant used in the final numbers, the exclusion set contains 47 features. This leaves 257 features in the strict non-style taxonomy used by the final narrative runs.

\begin{table*}[t]
\centering
\scriptsize
\resizebox{\textwidth}{!}{%
\begin{tabular}{p{0.12\textwidth} p{0.06\textwidth} p{0.18\textwidth} p{0.58\textwidth}}
\toprule
\textbf{Type} & \textbf{Count} & \textbf{Feature} & \textbf{Details} \\
\midrule

Categorical & 124 
& Dominant narrative person (\textit{Perspective})
& \textbf{Question:} What is the primary grammatical person used by the main narrator across the story? \\
& & 
& \textbf{Description:} Identify the default pronoun/verb alignment in narrative (not dialogue): `I/we', `you', `he/she/they'. If no dominant usage, mark `mixed\_no\_dominant'. \\
& & 
& \textbf{Answer choices:} `first\_person'; `second\_person'; `third\_person'; `mixed\_no\_dominant'. \\

\midrule

Ordinal & 59 
& Secondary character density (\textit{Agents})
& \textbf{Question:} How many distinct secondary/supporting characters appear in the story? \\
& & 
& \textbf{Description:} Include minor or functional characters; exclude major cast from \texttt{AGENT\_ID\_001}. Map counts to bins. \\
& & 
& \textbf{Answer choices:} `0--1'; `2--3'; `4--7'; `8+'. \\

\midrule

Scale & 45 
& Dyadic vs Group Scene Emphasis (\textit{Social Networks})
& \textbf{Question:} To what extent do key interactions occur in dyads versus groups? \\
& &
& \textbf{Description:} Evaluate major scenes. Rate 1 = mostly dyads, 3 = balanced, 5 = mostly groups. \\
& & 
& \textbf{Answer choices:} `1'; `2'; `3'; `4'; `5'. \\

\midrule

Binary & 44 
& Fourth-Wall Breaking (\textit{Situatedness})
& \textbf{Question:} Does the narration directly address the reader? \\
& & 
& \textbf{Description:} Look for explicit audience address (e.g., `dear reader'). Ignore in-world dialogue. \\
& & 
& \textbf{Answer choices:} `no'; `yes'. \\

\midrule

Multi-select & 32 
& Global Narrative Schema (\textit{Events})
& \textbf{Question:} Which high-level narrative schemas are central? \\
& & 
& \textbf{Description:} Identify structural patterns such as journey, mystery, transformation, ordeal, slice-of-life, trial, heist, or framed recollection. \\
& & 
& \textbf{Answer choices:} `quest/journey'; `investigation/mystery'; `transformation/redemption'; `siege/ordeal'; `slice\_of\_life'; `trial/test/game'; `heist/caper'; `frame\_confession/memoir'. \\

\bottomrule
\end{tabular}
}
\caption{Distribution of feature types in the feature taxonomy. An example feature for each type, with the question, description, and answer given, to illustrate the type of features searched for.}
\label{tab:feature_taxonomy_types}
\end{table*}


\section{Application of features.}
\label{app_sec:application}

\paragraph{Why not single shot application?}
We compared single-call application against aspect-based application over 12 story files. Aspect-based application produced far more complete feature vectors: average coverage rose from 68.4\% of features to 95.4\% of features. The remaining aspect-based misses were concentrated in a small number of harder-to-apply features, mostly in setting, whereas single-call application showed broad systematic dropout, especially in revelation and temporal-structure features. Later dimensions are omitted more often in the single-call set up, while the aspect-based variant preserves near-complete coverage by restricting each call to one narrative dimension.

\paragraph{Extractor reliability.}
We evaluate the reliability of Gemini~3 Flash (minimal thinking) as the production feature extractor via a repeated-measures design. This yields 300 total extraction outputs and, for fully present items, 600 within-item run pairs. We use the minimal thinking variant due to the much lower cost. Since we give full stories as part of the input context, having thinking on makes Gemini think for a very long time, even for the simple feature assignment tasks. On a smaller test of full vs. minimal thinking, the minimal thinking caused a 0.45\% reduction in performance on the binary task and an 11.61\% increase in performance on the 6-way attribution task. Aggregate inter-run agreement across the five runs is Krippendorff's $\alpha$ (nominal) = 0.90, mean pairwise Cohen's $\kappa$ = 0.89. 

\paragraph{Human validation against annotators.}\label{app_sec:human_validation}
We ran a small human validation study over 12 stories, 240 story-feature items per annotator. Two annotators completed the full set. We report agreement in that encoded representation. Against the model assignments, cohens $\kappa = 0.91$ for \emph{annotator 1} and cohens $\kappa = 0.7724$ for \emph{annotator 2}; the mean human--model cohens $\kappa$ is therefore 0.84 (see \autoref{tab:human_validation_summary}). Human--human encoded agreement on the same subset is cohens $\kappa = 0.74$.

\begin{table}[!t]
\centering
\scriptsize
\begin{tabular}{@{}lcc@{}}
\toprule
\textbf{Comparison} & \textbf{Encoded exact agreement (\%)} & \textbf{Cohen's $\kappa$} \\
\midrule
\emph{Annotator 1} vs model & 91.67 & 0.9056 \\
\emph{Annotator 2} vs model & 79.86 & 0.7724 \\
Mean human vs model & 85.76 & 0.8390 \\
Human vs human & 76.85 & 0.7385 \\
\bottomrule
\end{tabular}
\caption{Human validation on 240 story-features across 12 stories.}
\label{tab:human_validation_summary}
\end{table}
 
\section{Experimental Set Up}
\label{app_sec:experiments}

\paragraph{Model selection.}
We compared linear classifiers, random forests, and XGBoost on the validation split (100 prompts, 600 stories). XGBoost consistently outperformed the alternatives on both binary and multiclass tasks, and integrates cleanly with SHAP-based interpretability analyses \citep{lundberg2017shap,lundberg2020treeshap}, so we adopt it as the default classifier throughout.

\paragraph{Baseline setup.}
We evaluate four external baselines that operate on raw story text, without access to the LLM-extracted narrative features. 
\begin{enumerate}
    \item \textit{Stylometric baseline.} We extract 144 surface features from each story, including sentence-, word-, and paragraph-length statistics, document-level counts, vocabulary richness metrics, 100 function-word frequencies, punctuation rates, dialogue features, and readability indices.
    \item \textit{TF-IDF baseline.} We fit a unigram/bigram TF-IDF vectorizer. The resulting matrix is passed through the same XGBoost sweep and train/val/test protocol as the stylometric baseline.
    \item \textit{ModernBERT baseline.} We fine-tune ModernBERT-base \citep{warner2025modernbert} directly on raw story text using the same train/val/test split.
    \item \textit{Binoculars.} We run Binoculars in accuracy mode with all other settings left at their defaults.
\end{enumerate}

\paragraph{Feature encoding.}
For the binary task we set XGBoost's \texttt{scale\_pos\_weight}$=5$ to match the 5:1 AI-to-human class ratio. Multiclass uses uniform class weights. Features are encoded with one-hot columns for nominal and binary types and explicit integer encoding for ordinal and scale types. 


\paragraph{Evaluation protocol.}
The final split layout is 7,383 train / 1,405 val / 1,384 test prompts. We tune hyperparameters on the validation split, then retrain each final model on train+val (8,788 prompts; 52,707 stories) and report all final results on the held-out test set (1,384 prompts, 8,301 stories). For binary classification, we report macro-F1 and AUPRC.\footnote{With 5:1 class imbalance a trivial all-AI classifier achieves 83.3\% accuracy, so we omit accuracy from the binary table and rely on macro-F1 and AUPRC.} For 6-way attribution, we report macro-F1 and accuracy.

\paragraph{Feature role assignment.}
\label{app_sec:feature_roles}
A feature is marked \emph{important} if $\bar{s}_j$ exceeds the top-quartile threshold on mean absolute SHAP across features. Among important features, $j$ is labeled \emph{stable-important} iff (i) $\mathrm{stab}_j$ is at least the median stability score among important features and (ii) its mean SHAP exceeds the 95th-percentile null-label baseline with permutation $p \leq 0.10$; otherwise it is \emph{unstable-important}. Non-important features with above-median stability are labeled \emph{stable-weak}, and the remainder are \emph{noise}.

 \textbf{Core} features come from the \emph{binary} SHAP analysis and must satisfy all of the following: \emph{stable-important} quadrant membership, null significance, $\mathrm{stab}_j \geq 0.55$, $\mathrm{top25}_j \geq 0.60$, absolute human--AI mean-value gap at least $0.20$, and cross-model AI spread at most $0.35$.

\section{Dimension-level ablations}
\label{app_sec:dimension_ablation}

\emph{Dimension-only} retrains the classifier on a single NarraBench dimension; \emph{leave-one-out} retrains it on all narrative features except that dimension's, both under the protocol of \S\ref{app_sec:experiments}. No dimension is individually sufficient (best: \emph{agents} at 80.2\% binary macro-F1) and none is individually necessary (worst removal: \emph{agents} at $-$1.2), so the signal is redundant across correlated dimensions. This is expected, since AI-generated fiction has a recognizable style: e.g., more linear plots, tighter causal structure, more explicit thematic framing, and less diverse character and social structure often co-occur.

\begin{table}[!t]
\centering
\small
\begin{tabular}{@{}lcc c lcc@{}}
\toprule
\multicolumn{3}{c}{\textit{Dimension only}} & & \multicolumn{3}{c}{\textit{Leave-one-dimension-out}} \\
\cmidrule(r){1-3} \cmidrule(l){5-7}
\textbf{Dimension} & \textbf{Binary} & \textbf{6-way} & & \textbf{Dimension} & \textbf{Binary} & \textbf{6-way} \\
\midrule
Agents             & 80.2 & 46.0 & & Agents             & 92.0 ($-$1.2) & 66.3 ($-$2.1) \\
Situatedness       & 77.3 & 41.3 & & Situatedness       & 92.3 ($-$0.8) & 66.1 ($-$2.3) \\
Plot               & 74.5 & 40.5 & & Perspective        & 92.6 ($-$0.6) & 67.3 ($-$1.2) \\
Perspective        & 71.3 & 32.3 & & Setting            & 92.8 ($-$0.4) & 66.8 ($-$1.7) \\
Setting            & 70.6 & 38.4 & & Events             & 92.8 ($-$0.4) & 67.4 ($-$1.0) \\
Social networks    & 67.9 & 38.3 & & Plot               & 92.8 ($-$0.4) & 67.5 ($-$0.9) \\
Events             & 67.8 & 37.4 & & Temporal structure & 93.0 ($-$0.2) & 65.4 ($-$3.1) \\
Revelation         & 67.2 & 33.2 & & Social networks    & 93.0 ($-$0.2) & 67.8 ($-$0.6) \\
Temporal structure & 62.0 & 37.7 & & Revelation         & 93.5 ($+$0.3) & 67.7 ($-$0.7) \\
\midrule
\textit{All narrative} & \textit{93.2} & \textit{68.4} & & \textit{All narrative} & \textit{93.2} & \textit{68.4} \\
\bottomrule
\end{tabular}
\caption{Per-dimension ablations of the narrative model, macro-F1 (\%). Left: trained on one NarraBench dimension only. Right: trained on all narrative features except that dimension, with the change relative to the full narrative model in parentheses.}
\label{tab:dimension_ablation}
\end{table}

 \section{Memorization-Risk in story generation}
\label{app_sec:memorization}

Because the human corpus includes stories that are likely to be widely available on the public web and in Books3-adjacent collections, we run a conservative contamination-risk audit over the reverse-engineered outputs. Memorization could lead to the inclusion of style atypical to AI=writing.
We flag \emph{exact overlap} when the pair shares at least one exact 13-gram beyond split-matched shuffled-human controls, as done in prior work \citep{brown2020gpt3}, and \emph{near-verbatim overlap} when paired 8-gram coverage is at least 5\%, the pair shares at least four distinct 8-grams, and the longest exact common span is at least 30 tokens, again beyond shuffled-human controls. We measure output overlap rather than formal training-set membership, so we treat it as a sensitivity analysis rather than proof that a model saw a specific story during pretraining. Risk rates in \autoref{tab:memorization_audit_rates_full}.

\begin{table}[ht]
\centering
\scriptsize
\begin{tabular}{@{}lcc@{}}
\toprule
\textbf{Model} & \textbf{Exact 13-gram} & \textbf{Near-verbatim} \\
\midrule
GPT-5.4 & 64 (0.63\%) & 7 (0.07\%) \\
DeepSeek V3.2 & 58 (0.57\%) & 13 (0.13\%) \\
Kimi K2.5 & 44 (0.43\%) & 7 (0.07\%) \\
Gemini 3 Flash & 119 (1.17\%) & 32 (0.32\%) \\
Claude Sonnet 4.6 & 67 (0.66\%) & 14 (0.14\%) \\
\bottomrule
\end{tabular}
\caption{Memorization-risk rates by model.}
\label{tab:memorization_audit_rates_full}
\end{table}

The memorization screen flags 352 of 50,672 evaluated generated-story comparisons (0.70\%) as high risk. Under the conservative prompt-level ablation rule, this removes 169 of 10,139 audited prompts (1.67\%). The strongest cases are concentrated in canonical or heavily reprinted texts such as \emph{The Yellow Wallpaper}, \emph{The Legend of Sleepy Hollow}, \emph{Heart of Darkness}, \emph{The Call of Cthulhu}, and other Poe, Doyle, Wilde, and Dickens stories. 

\begin{table}[!t]
\centering
\scriptsize
\begin{tabular}{@{}lccc@{}}
\toprule
\textbf{Variant} & \textbf{Original test macro-F1} & \textbf{Low-risk test macro-F1} & \textbf{$\Delta$} \\
\midrule
Narrative + Style (binary) & 96.00 & 96.06 & +0.06 \\
Narrative + Style (multiclass) & 77.29 & 77.56 & +0.27 \\
Narrative (binary) & 93.18 & 93.23 & +0.05 \\
Narrative (multiclass) & 68.42 & 68.68 & +0.26 \\
\bottomrule
\end{tabular}
\caption{Sensitivity of the final detectability results after dropping all high-risk prompts.}
\label{tab:memorization_detectability_ablation}
\end{table}

Dropping the high-risk prompts leaves the main detectability conclusions essentially unchanged. In the filtered rerun, the Narrative + Style model improves slightly from 96.00 to 96.06 macro-F1 in binary detection and from 77.29 to 77.56 macro-F1 in 6-way attribution; the Narrative-only model shifts from 93.18 to 93.23 in binary and from 68.42 to 68.68 in multiclass (see \autoref{tab:memorization_detectability_ablation}).

\section{Additional Analysis}
\label{app_sec:length_confound}
\paragraph{Does length matter for narrative detection?} Human stories are substantially longer than AI stories on the held-out final test split (human mean / median $= 6418 / 4973$ words; AI mean / median $= 4523 / 3355$). To test whether binary detectability is primarily a length artifact, we run two additional analyses. We freeze the final binary classifiers from \autoref{sec:detectability} and re-evaluate them on a decile-stratified length-matched test subset with 2,754 stories (1,377 human, 1,377 AI) and nearly identical medians (human 4,973 vs.\ AI 4,951 words). The Narrative model is unchanged (93.2\% macro-F1 before and after matching), indicating that the main separation is not driven by the trivial human-longer / AI-shorter length contrast. The Style Only model is similarly stable, moving from 85.8\% to 86.8\% macro-F1 after matching. Because that matched-subset comparison alone does not establish perfect length invariance, we also evaluate the same frozen models separately on short, medium, and long bands defined by human-story tertiles within the matched subset. Narrative macro-F1 remains 91.6 / 94.3 / 93.7 across these bins, Narrative + Style remains 95.0 / 96.2 / 95.9, and Style Only remains 86.4 / 86.5 / 87.4, showing little variation with length overall. 

\begin{table}[!t]
\centering
\scriptsize
\begin{tabular}{@{}llcccc@{}}
\toprule
\textbf{Model} & \textbf{Setting} & \textbf{Macro-F1 (\%)} & \textbf{Human-F1} & \textbf{Bal.\ Acc.} & \textbf{AUPRC} \\
\midrule
Length only & Full test & 55.9 & 0.330 & 0.597 & 0.261 \\
Style Only & Full test & 85.8 & 0.766 & 0.871 & 0.867 \\
Style Only & Length-matched test & 86.8 & 0.859 & 0.869 & 0.956 \\
Narrative & Full test & 93.2 & 0.886 & 0.932 & 0.959 \\
Narrative & Length-matched test & 93.2 & 0.929 & 0.932 & 0.990 \\
Narrative + Style & Full test & 96.0 & 0.933 & 0.956 & 0.982 \\
Narrative + Style & Length-matched test & 95.7 & 0.956 & 0.957 & 0.996 \\
\bottomrule
\end{tabular}
\caption{Held-out final-test length-confound audit. ``Length only'' is a logistic regression trained on train+val story word counts alone. ``Length-matched test'' reuses the frozen final binary models but evaluates them on a 2,754-story subset (1,377 human, 1,377 AI) formed by stratifying AI stories into human-derived length deciles and sampling without replacement.}
\label{tab:length_confound_final}
\end{table}

\begin{wraptable}{r}{0.55\columnwidth}
\vspace{-8pt}
\centering
\resizebox{\linewidth}{!}{%
\begin{tabular}{@{}lcccccc@{}}
\toprule
& \textbf{Hum.} & \textbf{Cla.} & \textbf{GPT} & \textbf{Gem.} & \textbf{DS} & \textbf{Kimi} \\
\midrule
Narr.\ + Style & 0.93 & 0.89 & 0.82 & 0.67 & 0.66 & 0.67 \\
Narrative      & 0.89 & 0.77 & 0.73 & 0.60 & 0.57 & 0.55 \\
\bottomrule
\end{tabular}}
\caption{Per-class F1 for 6-way attribution. Human writing is easiest to classify, while Gemini, DeepSeek, and Kimi have the lowest performance.}
\label{tab:per_class_f1}
\vspace{-8pt}
\end{wraptable}

\paragraph{Does topic matter for detection?} We also examined whether performance varies by story topic. We prompted Qwen3.5-9B \citep{qwen35} to give a Book Industry Standards and Communications (BISAC) code \citep{bisg_bisac} given a story text, and apply this over the test set. Stories are categorized into  12 topics but only 6 have at least 20 test prompts. For the Narrative model, binary macro-F1 ranges from 90.0\% (mystery/detective) to 96.2\% (historical), while 6-way macro-F1 ranges from 65.8\% (mystery/detective) to 70.8\% (science fiction). 

To test whether mean detection performance differs by topic, we computed prompt-level test accuracy within each prompt's majority topic and ran a Kruskal--Wallis omnibus test over the six topics with at least 20 test prompts (literary, science fiction, horror, action/adventure, fantasy, mystery/detective). We find no significant topic-wise differences: $H{=}4.69$, $p{=}0.46$. So topic shifts the point estimates somewhat, especially for multiclass attribution, but we do not find strong evidence that detectability changes systematically by topic. 

\section{Per-story rarity analysis.}
\label{app_sec:rarity_analysis}

The group-level centroid and dispersion statistics above characterize sources regions but do not directly measure whether individual human stories are more structurally unusual. To operationalize the originality definition from \S\ref{sec:intro}, we compute per-story rarity as the mean Euclidean distance to a story's $k=25$ nearest neighbors in the pooled train+val narrative space. Rarity percentiles are computed against the train+val rarity distribution: a story at the 90th percentile sits in a sparser region of narrative space than 90\% of all train+val stories. All effect sizes here are computed over individual stories, not per-prompt summaries.

\begin{figure}[htpb]
\centering
\includegraphics[width=\linewidth]{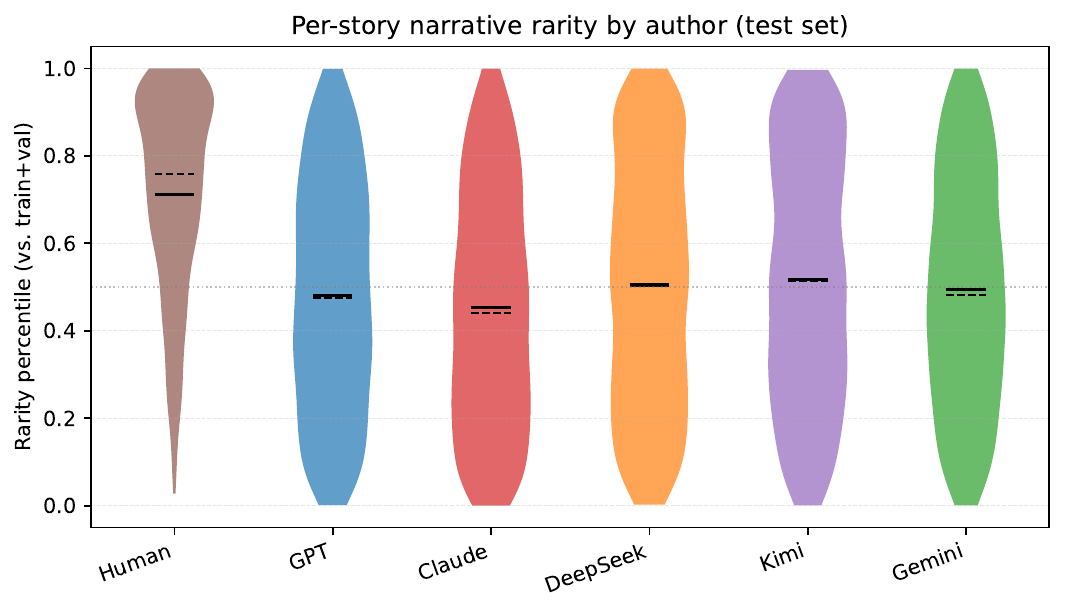}
\caption{Per-story narrative rarity percentiles by sources on the held-out test set. Solid lines show means; dashed lines show medians. Human stories are shifted toward higher rarity but all distributions overlap substantially.}
\label{fig:rarity_violin}
\end{figure}

\autoref{fig:rarity_violin} shows the per-sources rarity distributions on the held-out test set. Human stories are shifted toward higher rarity (mean percentile 0.71 vs.\ 0.49 for AI, AUC\,=\,0.73). Using the standard sample-size-weighted pooled SD over the 1{,}377 human and 6{,}885 AI test stories gives Cohen's $d$\,=\,0.83, with sample SDs of 0.227 for human stories and 0.274 for pooled AI stories and a pooled SD of 0.267.

At the prompt level, the human story is rarer than all five AI alternatives 57.8\% of the time (mean margin\,=\,0.96 raw distance units). The rarest test-set tail is mixed: the top 1\% contains 42 human and 41 AI stories; the top 5\% contains 180 human vs.\ 234 AI; the top 10\% contains 340 human vs.\ 487 AI stories across five models (\autoref{tab:rarity_tail}). 

\begin{table}[t]
\centering
\small
\begin{tabular}{@{}lrrrrrr@{}}
\toprule
\textbf{Tail} & \textbf{Human} & \textbf{Claude} & \textbf{GPT} & \textbf{DeepSeek} & \textbf{Kimi} & \textbf{Gemini} \\
\midrule
Top 1\%  & 42  & 5  & 5  & 11 & 6  & 14 \\
Top 5\%  & 180 & 33 & 34 & 58 & 64 & 45 \\
Top 10\% & 340 & 66 & 71 & 127 & 135 & 88 \\
\bottomrule
\end{tabular}
\caption{Composition of the rarest test-set stories by sources. Counts are out of 8,262 test stories (1,377 per sources). Human stories are overrepresented in all tails relative to a $\frac{1}{6}$ baseline, but AI stories are present throughout.}
\label{tab:rarity_tail}
\end{table}

\section{Core and Fingerprint Features}
\label{app_sec:feature_catalogs}


\autoref{tab:core_ai_features} lists the 20 features that reliably characterize AI writing in the binary task, ranked by a core score computed as mean SHAP multiplied by stability score and by one plus the absolute human--AI gap for the qualifying encoded column.

\begin{table*}[htbp]
\centering
\scriptsize
\begin{tabular}{@{}r p{0.20\textwidth} p{0.24\textwidth} l l p{0.25\textwidth}@{}}
\toprule
\textbf{\#} & \textbf{Feature} & \textbf{Question} & \textbf{Dim} & \textbf{Type} & \textbf{Response Options} \\
\midrule
1 & Thematic Explicitness and Moralizing & How explicitly does the story articulate its themes or morals? & SIT & scale & 1--5 \\
2 & Dominant Emotional Expression $\rightarrow$ embodied & How are characters' emotions most commonly conveyed? & AGENT & cat & explicit labels, embodied metaphors, behavioral cues, ambiguous \\
3 & Thematic Unity & To what extent do subplots and flourishes serve a central thematic concern? & PLT & scale & 1--5 \\
4 & Dominant Sensory Modalities $\rightarrow$ olfactory & Which sensory modalities does the story most frequently engage? & SET & multi & visual, auditory, olfactory, tactile, gustatory, kinesthetic \\
5 & Character Introduction $\rightarrow$ external description & What narrative device primarily introduces the central character? & AGENT & cat & external\_desc, in-action, in-dialogue, inner\_thought, others\_reports \\
6 & Setting as Psychological Mirror & To what degree does physical environment mirror characters' inner states? & SET & scale & 1--5 \\
7 & Continuity of Main Causal Chain & How continuous is the single causal chain from inciting incident to ending? & EVT & scale & 1--5 \\
8 & Sensory Density & How dense is sensory description across the narrative? & SET & scale & minimal--lush \\
9 & Agency in Resolution $\rightarrow$ protagonist choice & Is resolution driven by protagonist's choices or external events? & PLT & cat & protagonist\_choice, mixed, external\_fate \\
10 & Narratorial Thematic Commentary $\rightarrow$ yes & Does the narrator explicitly comment on themes beyond characters' perspectives? & SIT & binary & no, yes \\
11 & Opening Spatial Grounding & How clearly does the opening ground the reader in a specific physical setting? & SET & ord & none/vague, minimal, clear local, clear local+global \\
12 & Dialogue Function $\rightarrow$ philosophical debate & What main functions does dialogue serve? & PER & multi & advance plot, reveal character, worldbuilding, philosophical, comic \\
13 & Spatial Granularity Level & How fine-grained is the story's depiction of physical space? & SET & ord & very\_low--high \\
14 & Subplot Integration $\rightarrow$ no subplots & How directly do subplots echo the central theme? & PLT & cat & no\_subplots, thematically\_parallel, contrasting, independent \\
15 & Moral / Philosophical Weighting & How heavily does the story foreground moral or philosophical questions? & SIT & scale & 1--5 \\
16 & Reference Explicitness $\rightarrow$ implicit echoes & Are intertextual gestures primarily explicit or diffuse? & SIT & cat & none, explicit named, implicit echoes, balanced mix \\
17 & Environmental and Ecological Emphasis & How prominent is the natural environment or ecology in the narrative? & SET & scale & 1--5 \\
18 & Mode of Resolution $\rightarrow$ internal understanding & Is the main event chain resolved through internal acceptance or external action? & EVT & cat & resolved externally, resolved internally, unresolved \\
19 & Pre-Threat Character Investment & How much does the story build investment before major jeopardy? & REV & scale & 1--5 \\
20 & Depth of Interior Access & How deep into characters' inner life does narration go? & PER & scale & 1--5 \\
\bottomrule
\end{tabular}
\caption{Core AI-characterizing features (20). Dim = NarraBench dimension prefix; Type: scale = 1--5 Likert, cat = categorical, ord = ordinal, multi = multi-select, binary = yes/no. Arrow ($\rightarrow$) indicates the specific option value that is elevated for AI.}
\label{tab:core_ai_features}
\end{table*}

\autoref{tab:core_human_features} lists the 13 features that reliably characterize human writing, ranked by the same core score. 

\begin{table*}[!t]
\centering
\scriptsize
\begin{tabular}{@{}r p{0.20\textwidth} p{0.24\textwidth} l l p{0.25\textwidth}@{}}
\toprule
\textbf{\#} & \textbf{Feature} & \textbf{Question} & \textbf{Dim} & \textbf{Type} & \textbf{Response Options} \\
\midrule
1 & Intertextual Strategy Types $\rightarrow$ explicit named reference & What kinds of intertextual engagement does the story employ? & SIT & multi & explicit named, retelling, pastiche, myth/religion, self referential \\
2 & Frequency of Direct Reader Address & How often does the text directly address the reader? & PER & ord & never, occasional asides, frequent/structural \\
3 & Reference Explicitness $\rightarrow$ balanced mix & Are intertextual gestures explicit or diffuse? & SIT & cat & none, explicit named, implicit echoes, balanced mix \\
4 & Depth of Recontextualization After Surprise & How extensively does a revelation force reinterpretation of earlier scenes? & REV & scale & 1 (none)--5 (complete re-reading) \\
5 & Dialogue-to-Narration Proportion & What proportion of text is direct dialogue vs.\ narration? & PER & scale & 1 (no dialogue)--5 (dialogue dominates) \\
6 & Fourth-Wall Permeability & To what extent does the story break the boundary between story-world and reader? & SIT & ord & 1 (no breaking)--4 (radical violations) \\
7 & Subplot Integration $\rightarrow$ thematically parallel & How directly do subplots echo the central theme? & PLT & cat & no subplots, thematically parallel, contrasting, independent \\
8 & Degree of Chronological Discontinuity & How often does the narrative jump across time? & TMP & scale & 1--5 \\
9 & Location Variety Scope & How many distinct physical locales does the story inhabit? & SET & ord & single--multiworld \\
10 & Anachrony Intensity & How heavily does the narrative rely on flashbacks or flash-forwards? & TMP & scale & 1 (absent)--5 (dominant anachronic) \\
11 & Moral Polarity Toward Protagonist $\rightarrow$ ambivalent & Does the narrative frame the protagonist's choices as morally clear or ambiguous? & PLT & cat & clearly positive, ambivalent/mixed, clearly negative \\
12 & Dominant Emotional Expression $\rightarrow$ explicit labels & How are characters' emotions most commonly conveyed? & AGENT & cat & explicit labels, embodied metaphors, behavioral cues, ambiguous \\
13 & Nonlinear Framing for Delayed Disclosure & To what extent does the story use time jumps to stage revelations? & REV & scale & 1 (linear)--5 (heavily fragmented) \\
\bottomrule
\end{tabular}
\caption{Core human-characterizing features (13), Arrow ($\rightarrow$) indicates the specific option value elevated for human authors. Abbreviations as in \autoref{tab:core_ai_features}.}
\label{tab:core_human_features}
\end{table*}

\autoref{tab:core_features_themed} presents all 30 core features organized by interpretive theme, with human and AI mean values and gaps. This table is referenced from \autoref{sec:detectability}.

\begin{table*}[!t]
\centering
\small
\begin{tabular}{@{}p{0.50\textwidth} r r r@{}}
\toprule
\textbf{Feature} & \textbf{Human} & \textbf{AI} & \textbf{Gap} \\
\midrule
\multicolumn{4}{l}{\textit{AI-elevated: Thematic over-determination}} \\
\quad Thematic Explicitness \& Moralizing \tmark{s} & 3.28 & 3.94 & $-$0.65 \\
\quad Moral / Philosophical Weighting \tmark{s} & 3.26 & 3.68 & $-$0.42 \\
\quad Thematic Unity \tmark{s} & 4.41 & 4.74 & $-$0.33 \\
\quad Narratorial Thematic Commentary $\rightarrow$ yes & 52\% & 77\% & $-$25 \\
\quad Dialogue Function $\rightarrow$ philosophical debate & 34\% & 59\% & $-$25 \\
\quad Reference Explicitness $\rightarrow$ implicit echoes & 50\% & 72\% & $-$22 \\
\midrule
\multicolumn{4}{l}{\textit{AI-elevated: Sensory \& embodied performativity}} \\
\quad Emotional Expression $\rightarrow$ embodied & 38\% & 81\% & $-$42 \\
\quad Setting as Psychological Mirror \tmark{s} & 3.58 & 4.07 & $-$0.49 \\
\quad Environmental \& Ecological Emphasis \tmark{s} & 2.83 & 3.21 & $-$0.38 \\
\quad Sensory Modalities $\rightarrow$ olfactory & 57\% & 82\% & $-$26 \\
\quad Sensory Density \tmark{s} & 3.66 & 3.93 & $-$0.26 \\
\quad Depth of Interior Access \tmark{s} & 3.67 & 3.93 & $-$0.26 \\
\midrule
\multicolumn{4}{l}{\textit{AI-elevated: Structural streamlining}} \\
\quad Causal Chain Continuity \tmark{s} & 3.92 & 4.20 & $-$0.28 \\
\quad Spatial Granularity \tmark{o} & 2.27 & 2.53 & $-$0.26 \\
\quad Agency in Resolution $\rightarrow$ protagonist choice & 46\% & 69\% & $-$23 \\
\quad Character Introduction $\rightarrow$ external description & 30\% & 52\% & $-$22 \\
\quad Subplot Integration $\rightarrow$ no subplots & 57\% & 79\% & $-$22 \\
\quad Resolution Mode $\rightarrow$ internal understanding & 27\% & 47\% & $-$21 \\
\quad Opening Spatial Grounding \tmark{o} & 2.12 & 2.33 & $-$0.20 \\
\quad Pre-Threat Character Investment \tmark{s} & 2.76 & 2.99 & $-$0.23 \\
\midrule
\multicolumn{4}{l}{\textit{Human-elevated: Intertextual richness}} \\
\quad Intertextual Strategy $\rightarrow$ explicit named reference & 47\% & 24\% & +23 \\
\quad Reference Explicitness $\rightarrow$ balanced mix & 37\% & 16\% & +21 \\
\midrule
\multicolumn{4}{l}{\textit{Human-elevated: Reader engagement}} \\
\quad Fourth-Wall Permeability \tmark{o} & 0.67 & 0.39 & +0.28 \\
\quad Direct Reader Address \tmark{o} & 0.28 & 0.07 & +0.21 \\
\midrule
\multicolumn{4}{l}{\textit{Human-elevated: Temporal complexity}} \\
\quad Depth of Recontextualization After Surprise \tmark{s} & 3.28 & 2.95 & +0.34 \\
\quad Chronological Discontinuity \tmark{s} & 2.40 & 2.12 & +0.28 \\
\quad Nonlinear Framing for Delayed Disclosure \tmark{s} & 1.96 & 1.68 & +0.28 \\
\quad Anachrony Intensity \tmark{s} & 2.58 & 2.31 & +0.27 \\
\midrule
\multicolumn{4}{l}{\textit{Human-elevated: Narrative diversity}} \\
\quad Location Variety Scope \tmark{o} & 1.34 & 1.08 & +0.26 \\
\quad Dialogue-to-Narration Proportion \tmark{s} & 2.95 & 2.70 & +0.24 \\
\quad Subplot Integration $\rightarrow$ thematically parallel & 42\% & 21\% & +22 \\
\quad Moral Polarity $\rightarrow$ ambivalent/mixed & 59\% & 38\% & +21 \\
\quad Emotional Expression $\rightarrow$ explicit labels & 29\% & 8\% & +21 \\
\bottomrule
\end{tabular}
\caption{Core features separating human from AI writing, grouped by theme. Features marked \tmark{s} are 1--5 Likert scales (values are means); \tmark{o} are ordinal scales (values are means over integer codes); $\rightarrow$ indicates a specific categorical/binary option (values are prevalence \%). Gap = Human $-$ AI; negative gaps indicate AI-elevated features. AI column averages across all five models.}
\label{tab:core_features_themed}
\end{table*}

The number of fingerprints varies widely across sources (Human: 32, Claude: 26, GPT: 11, Gemini: 11, DeepSeek: 7, Kimi: 3). \autoref{tab:fingerprint_summary} provides a compact summary.

\begin{table*}[!t]
\centering
\scriptsize
\begin{tabular}{@{}l r p{0.38\textwidth} l r r@{}}
\toprule
\textbf{Author} & \textbf{\#} & \textbf{Feature} & \textbf{Dim} & \textbf{SHAP} & \textbf{Uniq.} \\
\midrule
\multirow{5}{*}{\textbf{Human}}
& 1 & Character introduction $\rightarrow$ in-dialogue & AGENT & 0.110 & 21.4 \\
& 2 & Breadth of focalization $\rightarrow$ single focal & PER & 0.083 & 15.7 \\
& 3 & Narrator address mode $\rightarrow$ no direct address & SIT & 0.069 & 12.6 \\
& 4 & Overall revelation pacing $\rightarrow$ back-loaded & REV & 0.094 & 7.7 \\
& 5 & Literary ambition $\rightarrow$ crossover genre & SIT & 0.116 & 6.8 \\
\cmidrule(l){2-6}
& \multicolumn{5}{@{}l}{\textit{+ 27 more (visibility of withholding, atmospheric techniques, subplot density, naming, twist placement, \ldots)}} \\
\midrule
\multirow{5}{*}{\textbf{Claude}}
& 1 & Strength of event escalation & EVT & 0.402 & 22.4 \\
& 2 & Event-type diversity & EVT & 0.491 & 10.7 \\
& 3 & Ending temporal scope $\rightarrow$ epilogue/flashforward & TMP & 0.096 & 8.9 \\
& 4 & Dreams/visions as temporal distortion $\rightarrow$ no & TMP & 0.116 & 7.7 \\
& 5 & Setting mood $\rightarrow$ uncanny/haunted & SET & 0.059 & 4.6 \\
\cmidrule(l){2-6}
& \multicolumn{5}{@{}l}{\textit{+ 21 more (event density, conflict modality, relationship trajectory, heteroglossia, closure, \ldots)}} \\
\midrule
\multirow{5}{*}{\textbf{GPT}}
& 1 & Role of gossip and rumor $\rightarrow$ salient & SOC & 0.200 & 22.1 \\
& 2 & Narrator temporal distance $\rightarrow$ distant retrospective & TMP & 0.119 & 6.8 \\
& 3 & Reader expectation strategy $\rightarrow$ subverts & REV & 0.098 & 3.9 \\
& 4 & Iterative/habitual narration $\rightarrow$ no & TMP & 0.144 & 3.2 \\
& 5 & Reconciliation/forgiveness $\rightarrow$ partial/ambiguous & SOC & 0.066 & 2.6 \\
\cmidrule(l){2-6}
& \multicolumn{5}{@{}l}{\textit{+ 6 more (community salience, social emphasis, reader competence, individualization, psych.\ depth, \ldots)}} \\
\midrule
\multirow{5}{*}{\textbf{Gemini}}
& 1 & Protagonist social trajectory $\rightarrow$ expands & SOC & 0.058 & 5.0 \\
& 2 & Balance of speech $\rightarrow$ primarily direct & PER & 0.119 & 3.6 \\
& 3 & Global narrative schema $\rightarrow$ siege/ordeal & EVT & 0.037 & 3.5 \\
& 4 & Naming practice $\rightarrow$ named personal name & AGENT & 0.033 & 3.2 \\
& 5 & Global chronological structure $\rightarrow$ frequent flashbacks & TMP & 0.060 & 3.1 \\
\cmidrule(l){2-6}
& \multicolumn{5}{@{}l}{\textit{+ 6 more (secondary char.\ density, batch intro, authority stance, setting mood, community, closure)}} \\
\midrule
\multirow{7}{*}{\textbf{DeepSeek}}
& 1 & Narrator presence/visibility & PER & 0.153 & 4.1 \\
& 2 & Emotional expression $\rightarrow$ behavioral cues & AGENT & 0.069 & 3.6 \\
& 3 & Plot vs.\ atmosphere orientation & SIT & 0.117 & 2.9 \\
& 4 & Backstory placement $\rightarrow$ evenly interleaved & TMP & 0.096 & 2.7 \\
& 5 & Embedded storytelling scenes & SIT & 0.078 & 2.2 \\
\cmidrule(l){2-6}
& \multicolumn{5}{@{}l}{\textit{+ 2 more (seasons/cyclical time, \ldots)}} \\
\midrule
\multirow{3}{*}{\textbf{Kimi}}
& 1 & Character introduction $\rightarrow$ in-action event & AGENT & 0.163 & 3.7 \\
& 2 & Narrative entry frame $\rightarrow$ in medias res & PLT & 0.035 & 3.0 \\
& 3 & Explicit trait labeling $\rightarrow$ no & AGENT & 0.136 & 2.0 \\
\bottomrule
\end{tabular}
\caption{Per-sources fingerprint features, showing the top 5 (or all, if fewer) ranked by uniqueness ratio. SHAP = mean class SHAP importance; Uniq.\ = uniqueness ratio vs.\ next-best class.
}
\label{tab:fingerprint_summary}
\end{table*}

\section{Prompts}
\label{app_sec:prompts}

Given a source story, this prompt asks the model to produce a single concise writing prompt that preserves the story's key characters, setting, and thematic direction while leaving room for variation. Prompt is in \autoref{prompt:prompt_generation}.

\appendixpromptfigure{prompts_display/prompt_generation.md}{Prompt used to generate benchmark writing prompts from source stories}{prompt:prompt_generation}

 We do not use a single writing template; each story is generated from a standalone narrative prompt in the Books3-derived prompt set, and the same prompt text is given to all six sources. \autoref{prompt:story_generation_example} shows one representative validation prompt. 

\appendixpromptfigure{prompts_display/story_generation_example.md}{Representative benchmark story-generation prompt}{prompt:story_generation_example}

After story generation, each story is converted into a JSON template with a single extraction prompt. \autoref{prompt:template_extraction} shows the exact markdown template.

\appendixpromptfigure{prompts_display/template.md}{Prompt for NarraBench template extraction}{prompt:template_extraction}




\end{document}